\documentclass[lettersize,journal]{IEEEtran}
\usepackage{graphicx} 
\usepackage{algorithm}
\usepackage{algorithmic}
\usepackage{amsmath}
\usepackage{amssymb}

\usepackage{booktabs}
\usepackage{multirow}
\usepackage{xcolor}
\title{FedRSClip: Federated Learning for Remote Sensing Scene Classification Using Vision-Language Models}
\author{Hui Lin, Chao Zhang, Danfeng Hong, \IEEEmembership{Senior Member, IEEE}, Kexin Dong, and Congcong Wen\textsuperscript{\textdagger} \IEEEmembership{Member, IEEE}
\thanks{This work was supported in part by the National Key Research and Development Program of China (2021YFC3300500) and Beijing Nova Program (2024124). \textit{(Corresponding author: Congcong Wen)}.}
\thanks{Hui Lin, Chao Zhang, and Kexin Dong are with China Academy of Electronics and Information Technology, Beijing 100846, China. (e-mail: linhui@whu.edu.cn, sa615513@mail.ustc.edu.cn and kexindong1113@gmail.com)}
\thanks{D. Hong is with the Aerospace Information Research Institute, Chinese Academy of Sciences, 100094 Beijing, China, and also with the School of Electronic, Electrical and Communication Engineering, University of Chinese Academy of Sciences, 100049 Beijing, China. (e-mail: hongdf@aircas.ac.cn).}
\thanks{Congcong Wen is with the Department of Electrical and Computer Engineering, New York University Abu Dhabi, Abu Dhabi, UAE. (e-mail: wencc@nyu.edu).}}

\begin{document}

\maketitle

\begin{abstract}
Remote sensing image classification is essential for various applications, including agricultural monitoring, urban planning, and land use classification. However, remote sensing data is often distributed across multiple institutions, and due to privacy concerns and data-sharing restrictions, leveraging large-scale datasets in a centralized training framework is challenging. Federated learning offers a promising solution by enabling collaborative model training across distributed data sources without requiring data centralization. However, current Vision-Language Models (VLMs), which typically contain billions of parameters, pose significant communication challenges for traditional federated learning approaches based on model parameter updates, as they would incur substantial communication costs. In this paper, we propose FedRSCLIP, the first federated learning framework designed for remote sensing image classification based on a VLM, specifically CLIP. FedRSCLIP addresses the challenges of data heterogeneity and large-scale model transmission in federated environments by introducing Prompt Learning, which optimizes only a small set of tunable parameters. The framework introduces a dual-prompt mechanism, comprising Shared Prompts for global knowledge sharing and Private Prompts for client-specific adaptation. To maintain semantic coherence between shared and private prompts, we propose the Dual Prompt Alignment Constraint to balance global consistency and local adaptability across diverse client distributions. Additionally, to enhance cross-modal representation learning, we introduce the Cross-Modal Feature Alignment Constraint to align multimodal features between text and image prompts. To validate the effectiveness of our proposed model, we construct a Fed-RSIC dataset based on three existing remote sensing image classification datasets, specifically designed to simulate various federated learning configurations. Experimental results on the Fed-RSIC dataset demonstrate the effectiveness and superiority of FedRSCLIP in addressing the challenges of federated remote sensing image classification.
\end{abstract}

\section{Introduction}
Remote sensing image classification has emerged as a critical technique in diverse fields, such as agricultural monitoring~\cite{tian2023shape}, urban planning~\cite{han2024autoencoding}, land use classfication~\cite{rwanga2017accuracy}, and environmental forecasting~\cite{wen2019novel}. These applications rely on the ability to interpret vast amounts of data captured by satellites, drones, and aerial platforms to extract meaningful insights. The large-scale nature of remote sensing data, combined with the growing demand for timely and accurate analysis, underscores the importance of developing robust image classification models. However, due to the distributed nature of remote sensing data, which is often collected across various geographic locations and stored in different institutions, privacy concerns arise when centralizing data. In response to these concerns, federated learning~\cite{advances}  has emerged as a powerful approach, enabling collaborative learning across distributed datasets while preserving data privacy by keeping raw data local. This advancement not only addresses data privacy concerns but also enables widespread, cross-institutional insights that can benefit industries reliant on accurate, large-scale image analysis.

\begin{figure}[t]
    \centering
    \includegraphics[width=1.0\linewidth]{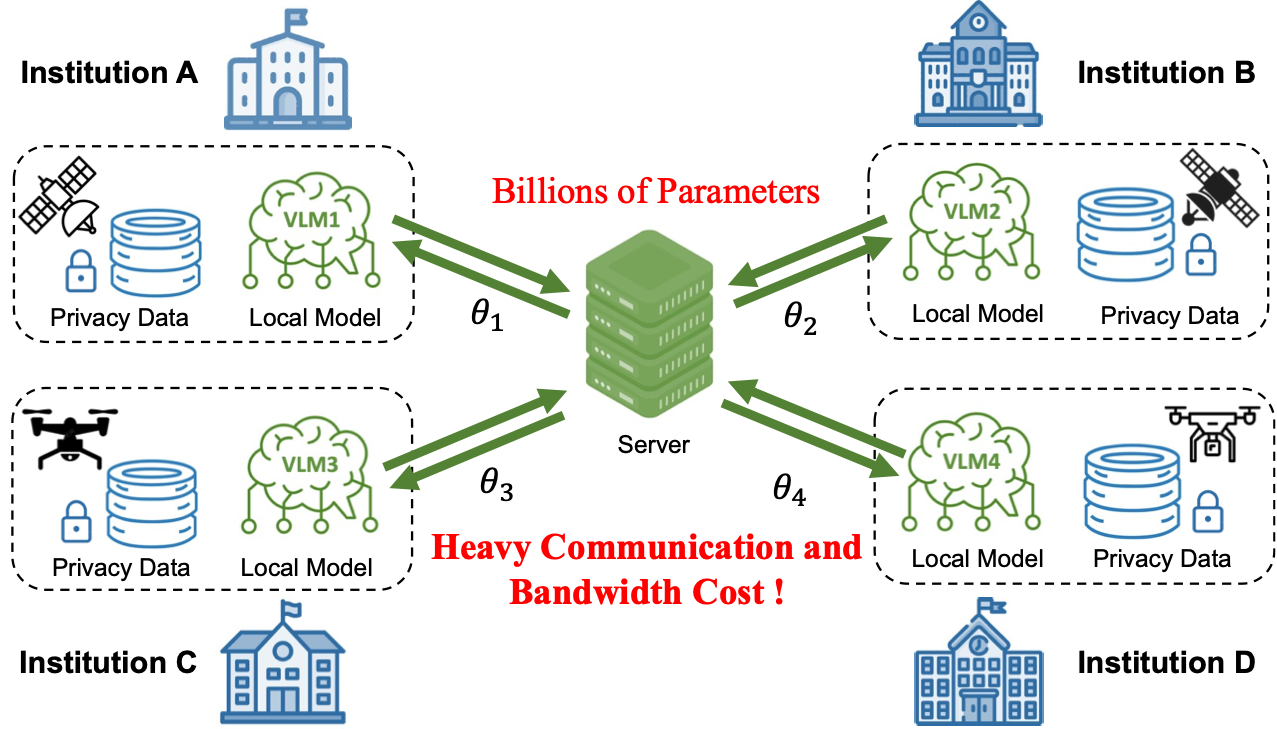}
    \caption{Illustration of Federated Learning Using Vision-Language Models (VLMs) in Remote Sensing Tasks. Remote sensing data is typically distributed across different institutions, with privacy concerns and data-sharing restrictions. Traditional federated learning involves uploading each client’s local model parameters to a central server for unified updates, which are then sent back to each client. However, for each client using VLM, transmitting billions of parameters each time leads to heavy communication and bandwidth costs. }
    \label{fig:intro}
\end{figure}

In recent years, federated learning has achieved notable success in several applications~\cite{fedavg,fedprox,fedalign}, leveraging deep learning models to enable decentralized training across clients. These models have demonstrated the potential to handle the non-iid (non-independent, non-identical) data distributions commonly seen in remote sensing. However, despite these successes, current federated learning approaches have largely focused on traditional deep learning models, which primarily learn from pixel-level image data without incorporating the rich contextual information that could be gained from other modalities. Vision-Language Models (VLMs)~\cite{zhu2023minigpt,dai2023instructblip,liu2024visual}, such as CLIP~\cite{clip}, offer a promising solution to this limitation. VLMs learn joint representations of images and text, mapping them into a shared feature space, enabling models to capture semantic relationships that go beyond visual features alone. By integrating VLMs into federated learning, we can potentially enhance model performance on remote sensing tasks, where both visual and textual descriptions, could significantly improve classification accuracy.

Despite recent successes in applying VLMs to remote sensing tasks~\cite{hu2023rsgpt,bazi2024rs,pang2024h2rsvlm,kuckreja2023geochat,lin2024rs}, adapting them to federated learning frameworks presents significant challenges. Traditional federated learning typically involves uploading each client’s local model parameters to a central server for unified updates, which are then sent back to each client. However, as illustrated in Fig.~\ref{fig:intro}, VLMs are large-scale models with millions or even billions of parameters, making it impractical to transmit the full model between clients and a central server due to substantial communication overhead. Additionally, the heterogeneity of data across clients, including variations in resolution, terrain, and spectral characteristics, further complicates the alignment of multimodal features within distributed settings. Thus, traditional federated learning algorithms, initially designed for smaller models, are not suited to handle the complexity and scale of VLMs. Therefore, an efficient strategy is essential to minimize communication load while maintaining VLM capabilities in federated setups.

To address these challenges, we propose FedRSCLIP, a novel \textbf{Fed}erated learning framework for \textbf{R}emote \textbf{S}ensing image classification based on \textbf{CLIP}, specifically designed to overcome the communication limitations of VLMs in federated learning. To the best of our knowledge, FedRSCLIP is the first federated learning framework to integrate VLMs into remote sensing image classification, leveraging both visual and textual information to enhance representation and classification. FedRSCLIP enables robust and generalized learning across distributed, non-iid datasets while preserving data privacy. Specifically, FedRSCLIP introduces Prompt Learning, which optimizes a small set of tunable parameters, significantly reducing communication overhead while maintaining adaptability to local data. To balance global consistency with local flexibility, FedRSCLIP employs a dual-prompt mechanism comprising Shared Prompts for global knowledge sharing and Private Prompts for client-specific customization. This mechanism ensures the model captures overarching patterns across all clients while allowing for fine-grained adaptations to diverse local data distributions. To maintain semantic consistency between shared and private prompts, we propose the Dual Prompt Alignment Constraint, which ensures that private prompts remain aligned with the global knowledge encapsulated by shared prompts, even as they adapt to local data. Additionally, we propose the Cross-Modal Feature Alignment Constraint, which aligns multimodal features across text and image prompts, facilitating more effective cross-modal representation learning and enhancing the overall coherence of the model's representations. Finally, we validate the effectiveness of FedRSCLIP on our newly constructed Fed-RSIC dataset, which is built upon three popular remote sensing image classification datasets: Optimal-31, UCMerced, and NWPU. Experimental results demonstrate that FedRSCLIP achieves state-of-the-art remote sensing image classification performance across various federated learning configurations. The contributions of this paper are summarized as follows:
 
\begin{itemize}
    \item We propose FedRSCLIP, the first framework to integrate Vision-Language Models into federated learning for remote sensing image classification. FedRSCLIP enhances representation and classification performance while optimizing communication efficiency, addressing challenges such as high communication costs and data heterogeneity.
    
    \item We introduce Prompt Learning for VLMs in federated learning, which optimizes a small set of tunable parameters instead of transmitting the entire model. Furthermore, we propose a dual-prompt mechanism comprising Shared Prompts for global knowledge sharing and Private Prompts for client-specific customization, enabling the model to balance global consistency and local flexibility.
    
    \item We develop two innovative constraints to enhance prompt alignment and representation learning in federated learning with VLMs. The first is the Dual Prompt Alignment Constraint, which ensures semantic consistency between shared and private prompts by aligning their representations during training. The second is the Cross-Modal Feature Alignment Constraint, which aligns multimodal features between text and image representations, enhancing the model's ability to capture relevant features and improving its classification performance.
    
    \item We construct the Fed-RSIC dataset by integrating three popular remote sensing image classification datasets. This dataset is specifically designed to simulate diverse federated learning scenarios, enabling comprehensive evaluation of the proposed framework. Experimental results demonstrate that FedRSCLIP achieves state-of-the-art performance in remote sensing image classification across various federated learning configurations.
\end{itemize}

\section{Related Work}
\subsection{Federated Learning}

Federated Learning has emerged as a key decentralized machine learning approach, preserving data privacy while allowing collaborative model training \cite{advances}. The foundational algorithm, FedAvg \cite{fedavg}, aggregates model updates from distributed clients without requiring local data sharing. While robust in handling non-IID data under synchronous updates, FedAvg's performance declines significantly when faced with heterogeneous data or evolving client datasets \cite{noniiddata, timeevolving1, timeevolving2}. Several methods have been proposed to address these challenges. For example, FedProx \cite{fedprox} introduces a proximal term to FedAvg, stabilizing local updates by constraining model divergence from the global model. Elastic aggregation \cite{elastic} adjusts update magnitudes based on parameter sensitivity, ensuring that the global model adapts appropriately to diverse client data. Moreover, approaches like FedSeg \cite{fedseg} modify cross-entropy loss to address class heterogeneity in specific tasks, such as semantic segmentation, while FedH2L \cite{fedh2l} and FedAlign \cite{fedalign} leverage distillation techniques to enhance generalization and reduce communication overhead by exchanging only selective information.

To further mitigate data heterogeneity, methods like FedFed \cite{fedfed} and FedOTP \cite{fedotp} employ feature distillation and prompt learning, enabling models to capture both global consensus and client-specific features. In contrast, aggregation-focused techniques such as FedAF \cite{fedaf} utilize client-condensed data to reduce client drift and accelerate convergence. In scenarios where client data evolve over time, methods like CFL \cite{cfl}, FedTHE \cite{fedthe}, and pFedEM \cite{pfedem} address temporal heterogeneity by adapting model architectures to dynamically changing distributions. Additionally, frameworks such as FBL \cite{timeevolving2} and FedRC \cite{fedrc} mitigate the forgetting of old classes and handle multiple types of distribution shifts through semantic compensation and robust clustering techniques, respectively.

\subsection{Federated Learning in Remote Sensing}
Federated learning has become increasingly relevant for remote sensing applications due to its ability to enable distributed data processing while maintaining data privacy. Various methods have been developed to address specific challenges in this domain. For instance, FedPM \cite{fedpm}, based on prototype matching, enhances object extraction performance in Deep Convolutional Neural Networks. Similarly, GeoFed \cite{geofed} targets semantic segmentation in earth observation, refining its objective function through Tail Regeneration and Essential Feature Mining strategies. Additionally, the architecture combining the deep memory connected neural network with the data-decoupled federated learning framework \cite{dmcn} facilitates image restoration while preserving privacy.

In the context of remote sensing image classification, recent advances in FL have focused on addressing challenges related to resource optimization, secure communication, and the handling of non-IID data. For instance, an adaptive model communication scheme \cite{adaptive} leverages deep Q-learning to optimize resource control in multi-access edge computing environments, using an epsilon-greedy strategy to allocate computation resources efficiently. To enhance security and efficiency, FLBIC-CUAV \cite{uav} integrates clustering, blockchain, and FL, utilizing beetle swarm optimization (BSO) for UAV clustering, blockchain for secure data transmission, and FL with a Residual Network model for cloud-based image classification. Similarly, the blockchain-empowered PPFL framework \cite{ppfl}, which employs the CKKS cryptosystem \cite{ckks}, enables satellite imagery owners to collaborate globally while ensuring data privacy and transparency. To improve classification performance across diverse data modalities, a multi-modal FL framework \cite{multimodal} associates images from different clients with various modalities, enhancing the robustness of the model. Additional strategies, such as feature-centric communication and pseudo-weight amalgamation, have been explored to improve the efficiency of model aggregation in FL \cite{leveragingfeature}. In the context of image generation for land use and cover classification, the integration of deep convolutional generative adversarial networks into FL frameworks \cite{dcgan} has proven effective, enabling client devices to generate high-quality images. To address the challenge of non-IID data across clients, transformer architectures have been introduced \cite{transformer}, particularly for multi-label classification tasks in RS. Additionally, FedDiff \cite{feddiff} offers a novel multi-modal diffusion-based FL framework, which combines dual-branch diffusion models for feature extraction with a lightweight communication module, ensuring efficient and private collaboration among clients. These advancements illustrate the growing sophistication of FL in RS, as it continues to tackle diverse challenges while maintaining data privacy and improving model performance.

\subsection{CLIP in Remote Sensing}
Contrastive Language-Image Pretraining (CLIP) \cite{clip} is an advanced vision-language model composed of a vision encoder and a text encoder, which generate vector representations for images and text, respectively. CLIP is trained using contrastive learning, where the model learns to associate correct image-text pairs while distinguishing them from incorrect ones. In recent years, the application of CLIP in remote sensing (RS) has attracted significant attention, as researchers have begun exploring how its vision-language capabilities can be adapted to the unique challenges of this domain.

Recently, several adaptations of CLIP have been developed for various RS tasks. For example, RS-CLIP \cite{rs-clip} combines contrastive vision-language pretraining with pseudo-labeling and curriculum learning to enhance semantic-aware visual representations and improve overall model performance. Similarly, ChangeCLIP \cite{changeclip}, designed for remote sensing change detection, modifies CLIP to extract bitemporal features and introduces a differential features compensation module to capture detailed semantic changes. This is further complemented by a vision-language-driven decoder to enhance image semantics. Additionally, PIR-CLIP \cite{pir-clip} applies prior instruction representation learning, pre-training on coarse-grained remote sensing data before fine-tuning on fine-grained data. It also introduces a cluster-wise attribution loss to reduce semantic confusion, further improving the model's ability to handle complex RS data. Moreover, SG-CLIP \cite{sg-clip} integrates geographic information with CLIP’s vision-language capabilities to boost species recognition accuracy, especially in few-shot learning scenarios. Similarly, GeoChat \cite{geochat}, built upon CLIP-ViT(L-14) \cite{clip} and fine-tuned with LLaVA-1.5 \cite{llava} using the LoRA \cite{lora} technique, extends CLIP's conversational abilities while enhancing its domain-specific knowledge for RS tasks. Furthermore, a methodology \cite{modalitygap} has been proposed to align RS imagery with the visual and textual modalities of CLIP through a two-stage process. This approach involves fine-tuning CLIP and performing cross-modal alignment, significantly improving performance in RS image classification and retrieval tasks.

\begin{figure*}
    \centering
    \includegraphics[width=1\linewidth]{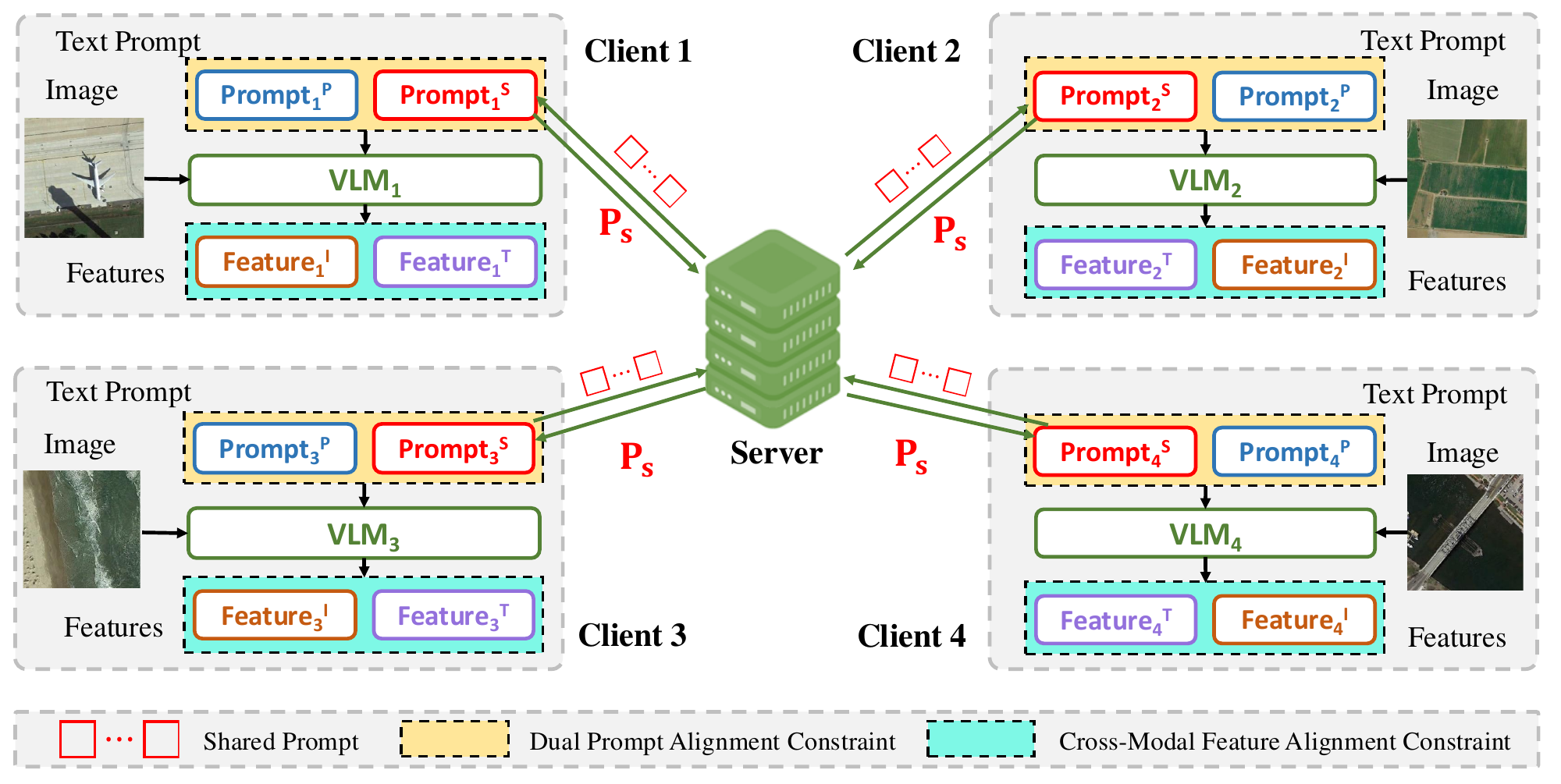}
    \caption{Illustration of FedRSCLIP's federated learning framework for remote sensing image classification using VLMs across multiple clients. Each client contains its own VLM, processing both image and text prompts. The framework utilizes a dual prompt mechanism, with \textit{Shared Prompts (Prompt\(^S\))} for global knowledge sharing across clients, and \textit{Private Prompts (Prompt\(^P\))} tailored to each client's unique data distribution. The \textit{Dual Prompt Alignment Constraint} (yellow rectangle with black dashed border) ensures alignment between shared and private prompts, while \textit{Cross-Modal Feature Alignment Constraint } (cyan rectangle with black dashed border) aligns textual and image features within each client to capture meaningful multimodal information. The server facilitates global updates by aggregating shared prompts (P\(_S\)) from all clients to improve the model's generalization capabilities across heterogeneous data environments.}
    \label{fig:overview}
\end{figure*}

\section{Dataset Creation}
Remote sensing datasets are inherently diverse, capturing variations in geographic regions, land-use types, spatial resolutions, and imaging conditions. However, most existing datasets are structured for centralized machine learning paradigms, lacking the characteristics needed to reflect the challenges of federated learning. To address this gap and better evaluate the performance of federated learning in remote sensing image classification tasks, we propose a novel dataset named Fed-RSIC, specifically designed for federated learning environments. Fed-RSIC is constructed by integrating three widely recognized remote sensing image classification datasets: Optimal-31, UCMerced, and NWPU-RESISC45. These datasets are carefully selected due to their diversity in land-use categories, image resolutions, and spatial characteristics, providing a robust foundation for simulating federated learning scenarios. To reflect their adaptation for federated learning, these datasets are restructured and renamed within the FedRSIC framework as Fed-Optimal, Fed-UCMerced, and Fed-NWPU, maintaining their inherent diversity while tailoring them to address federated learning challenges.

\subsection{Source Datasets}
\paragraph{Optimal-31} 
The OPTIMAL-31 dataset \cite{optimal-31}, collected from high-resolution Google Earth imagery, consists of 31 land-use classes, with a total of 1,860 images. Each class is represented by 60 images, all with a resolution of 256 × 256 pixels and a spatial resolution of 0.3 meters per pixel. This fine-grained spatial resolution allows the dataset to capture detailed ground features, making it suitable for various land-use classification and remote sensing tasks. The dataset covers a wide variety of categories, including both natural and man-made environments such as airports, baseball fields, basketball courts, churches, round farmland, dense housing areas, deserts, forests, golf courses, and meadows. 

\paragraph{UCMerced}
The UC Merced Land-Use dataset \cite{UCMerced} is a well-curated ground-truth image dataset, manually extracted from the USGS National Map Urban Area Imagery collection. It contains 2,100 RGB images, each measuring 256 × 256 pixels, with a spatial resolution of 0.3 meters per pixel, enabling fine-grained analysis of land-use patterns. The dataset is divided into 21 distinct land-use categories, with each category comprising 100 images. These categories cover a diverse range of environments, including agricultural fields, airplanes, beaches, buildings, chaparral, dense residential areas, and medium-density residential areas. 

\paragraph{NWPU}
The NWPU-RESISC45 dataset \cite{nwpu} is a large-scale benchmark designed for remote sensing image scene classification, developed by Northwestern Polytechnical University (NWPU). It contains 31,500 images, each with a resolution of 256 × 256 pixels. The spatial resolution of the images varies from 20 cm per pixel to over 30 meters per pixel, providing a diverse range of scales for analysis. The dataset is organized into 45 scene classes, with 700 images per class, featuring high intra-class diversity and inter-class similarity. These classes cover a wide array of environments, including airports, baseball diamonds, forests, harbors, freeways, overpasses, and ships.

\subsection{Federated Learning Simulation}
To simulate federated learning scenarios, the integrated datasets are partitioned into subsets corresponding to client configurations of 2, 5, 10, 15, 20, and 40 clients. These datasets, renamed as Fed-Optimal, Fed-UCMerced, and Fed-NWPU, are specifically tailored for federated learning experiments. For each configuration, the images within each dataset are evenly distributed among the specified number of clients. When the number of images per class is not perfectly divisible by the number of clients, the remaining images are distributed as evenly as possible across clients. This approach minimizes imbalance while ensuring fairness in data allocation. Before client-specific partitioning, the datasets are first split into training and testing sets. Specifically, 50\% of the images in Fed-Optimal and Fed-UCMerced are allocated for training, with the remaining 50\% reserved for testing. For the Fed-NWPU dataset, 20\% of the images are used for training, and the remaining 80\% are designated for testing. Table~\ref{tab:fedrsic} illustrates the image distribution per client for training and testing across each dataset under different client configurations. For instance, in the 40-client configuration for Fed-NWPU, each client receives approximately 158 training images and 630 testing images, with the remainder distributed among clients to maintain near-equal allocation. This partitioning strategy ensures that each client is assigned a balanced subset of data, preserving the inherent characteristics of the original datasets. By simulating federated learning conditions, the Fed-RSIC dataset provides a robust benchmark for evaluating federated learning models. The diversity of client configurations enables comprehensive exploration of various federated setups, allowing researchers to study the impact of client heterogeneity, data distribution, and scalability on model performance.

\begin{table}[h!]
\centering
\caption{Image distribution per client for training and testing across each dataset under different client configurations.}
\resizebox{0.48\textwidth}{!}{
\begin{tabular}{@{}c|cc|cc|cc@{}}
\toprule
\multirow{2}{*}{{\# Clients}} & \multicolumn{2}{c|}{{Fed-Optimal}} & \multicolumn{2}{c|}{{Fed-UCMerced}} & \multicolumn{2}{c}{{Fed-NWPU}} \\ 
                    & {Train} & {Test} & {Train} & {Test} & {Train} & {Test} \\ \midrule
2                  & 465             & 465           & 525             & 525           & 3,150           & 12,600         \\ 
5                  & 186             & 186           & 210             & 210           & 1,260           & 5,040          \\ 
10                 & 93              & 93            & 105             & 105           & 630             & 2,520          \\ 
15                 & 62              & 62            & 70              & 70            & 420             & 1,680          \\ 
20                 & 46              & 46            & 52              & 52            & 315             & 1,260          \\ 
40                 & 23              & 23            & 26              & 26            & 158             & 630            \\ \bottomrule
\end{tabular}}
\label{tab:fedrsic}
\end{table}

\section{Methods}

\subsection{Problem Statement}

In the task of federated learning for remote sensing image classification, there are $N$ clients distributed across different geographic locations, denoted as $\{\mathcal{C}_1, \mathcal{C}_2, \dots, \mathcal{C}_N\}$. Each client $\mathcal{C}_i$ possesses a local dataset of remote sensing images $\mathcal{D}_i = \{(I_j, y_j)\}_{j=1}^{m_i}$, where $I_j$ represents a remote sensing image, and $y_j \in \mathcal{Y}$ is the corresponding land cover class label. The goal is to collaboratively train a global classification model $f_{\theta}$ that can accurately classify images across all clients while adapting to the specific data distribution of each client. To achieve this goal, each client $\mathcal{C}_i$ is equipped with a local model $f_{\theta_i}$, which is trained to classify remote sensing images in the local dataset $\mathcal{D}_i$. For each image $I_j$, the local model computes the predicted class $\hat{y}_j = f_{\theta_i}(I_j)$ using the parameters $\theta_i$. The objective of the local model is to minimize its loss function on the local data:
\begin{equation}
\mathcal{L}_i(\theta_i) = \frac{1}{m_i} \sum_{j=1}^{m_i} \ell(f_{\theta_i}(I_j), y_j)
\end{equation}
where $\ell(\cdot)$ is the loss function used to measure the difference between the predicted values and the true labels (e.g., cross-entropy loss).

To leverage the strengths of each local model while maintaining data privacy, federated learning aggregates these locally optimized parameters into a global model. In federated learning, each client \( \mathcal{C}_i \) locally optimizes its model parameters \( \theta_i \) and periodically sends these parameters to a central server. The server collects the parameters \( \{\theta_i\}_{i=1}^N \) from all clients and aggregates them into global model parameters \( \theta \) using a weighted average:
\begin{equation}
\theta = \frac{1}{N} \sum_{i=1}^{N} w_i \theta_i
\end{equation}
where \( w_i = \frac{m_i}{m} \) represents the data weight of each client, and \( m = \sum_{i=1}^{N} m_i \) is the total number of samples across all clients. The aggregated global model \( f_{\theta} \) is then sent back to each client to update their local model parameters:
\begin{equation}
\theta_i \leftarrow \theta
\end{equation}
This process is repeated iteratively, with multiple rounds of model updates and parameter aggregation, ultimately forming a global model that generalizes well across all clients. The goal of federated learning is to maximize the classification performance of the global model \( f_{\theta} \) while preserving data privacy, ensuring that it maintains high accuracy on the data distributions present at each client.

\subsection{Overview}

In this paper, we propose FedRSCLIP, a federated learning framework for remote sensing image classification based on CLIP, designed to address the complexities of non-iid data across diverse clients. FedRSCLIP leverages CLIP’s ability to learn cross-modal representations, capturing fine-grained relationships between images and textual descriptions, which is crucial for remote sensing image classification tasks. To minimize the communication overhead associated with transmitting the large CLIP model across clients, FedRSCLIP incorporates Prompt Learning, optimizing only a small set of learnable prompt parameters. This approach enables efficient adaptation to local client data while significantly reducing communication costs. Furthermore, FedRSCLIP employs a dual prompt mechanism consisting of \textit{Shared Prompts} and \textit{Private Prompts} to achieve a balance between global consistency and local adaptability. Shared Prompts capture global features across all clients, facilitating knowledge sharing, while Private Prompts are customized to the unique data distribution of each client, enabling personalized model adjustments. To further ensure semantic consistency between Shared and Private Prompts, FedRSCLIP introduces the \textit{Dual Prompt Alignment Constraint}, which maintains the alignment of Private Prompts with global knowledge while allowing them to adapt to local data distributions. Additionally, to improve cross-modal alignment between textual and image features, FedRSCLIP incorporates the \textit{Cross-Modal Feature Alignment Constraint}, which enhances the model's ability to extract relevant features from both modalities. This constraint improves the coherence of multimodal representation learning, ensuring robust and accurate classification across federated settings. By integrating these techniques, FedRSCLIP effectively addresses the challenges posed by data heterogeneity in federated learning environments, resulting in improved global generalization and robust local adaptability for remote sensing image classification.

\subsection{CLIP and Prompt Learning}

Inspired by the recent success of VLMs in many vision tasks, we attempt to introduce VLMs into the federated learning framework to address the key challenges in remote sensing image classification. In our work, we employ the classical VLM model, \textit{CLIP}, as the classification model on each client. CLIP learns joint representations of images and text, mapping both into a shared embedding space, which allows it to capture fine-grained relationships between images and textual descriptions. This ability makes CLIP particularly well-suited for cross-modal information extraction and matching, especially in complex tasks such as remote sensing image classification. By utilizing textual prompts, CLIP can precisely identify different land cover classes, enhancing both classification accuracy and generalization across diverse remote sensing scenarios. However, one major limitation of the CLIP model lies in its large number of parameters. In federated learning, where each client must frequently exchange model parameters with the central server, transmitting the entire CLIP model would drastically increase communication costs. This problem becomes even more pronounced when involving multiple heterogeneous clients. Therefore, it is crucial to reduce communication overhead while maintaining the powerful representation capabilities of CLIP.

To tackle this issue, we introduce the \textit{Prompt Learning} method. Instead of transmitting the entire CLIP model, we optimize a small number of prompt parameters, allowing them to adapt to the local data on each client. This significantly reduces communication costs while enabling each client to quickly adapt to its local data, preserving the effectiveness of CLIP in different tasks. CLIP consists of two branches: the image branch, which contains rich information, and the text branch, which typically has weaker information. The textual prompts are often manually designed, such as "a photo of a [Class]". However, such fixed template prompts are overly broad and cannot sufficiently capture class-specific characteristics, limiting classification performance. Additionally, manually designing prompts is time-consuming and lacks generalization. Therefore, through Prompt Learning, we transform the text prompts into a learnable format, allowing the prompts to be automatically optimized based on the downstream task, replacing the need for hand-crafted prompts.

Specifically, Prompt Learning introduces $h$ learnable vectors into the text prompts. We place the class token ([CLASS]) in the middle of the sequence, and the prompt $s$ is structured as follows:
\begin{equation}
    s = [V]_1 [V]_2 \dots [V]_{\frac{h}{2}}[\text{CLASS}][V]_{\frac{h}{2} + 1} \dots [V]_h,
\end{equation}
where each $[V]_m$ ($m \in \{1, \dots, h\}$) is a vector with the same dimension as word embeddings. By forwarding the prompt $s$ to the text encoder $g(\cdot)$, we obtain prediction probability, which is computed as:
\begin{equation}
p(y = i | I) = \frac{\exp(\cos(g(s_i), f(I)) / \tau)}{\sum_{j=1}^{K} \exp(\cos(g(s_j), f(I)) / \tau)},
\end{equation}
where $s_i$ denotes the prompt for class $i$, where the class token [CLASS] is replaced by the corresponding word embedding vector(s) for class $i$. And $g(\cdot)$ is the text encoder, $f(\cdot)$ is the image encoder, $\cos(\cdot, \cdot)$ denotes the cosine similarity, $\tau$ is the temperature parameter for the softmax function, and $K$ is the total number of classes.

\subsection{Shared and Private Prompt Learning}

Considering that each client’s data has its own local distribution, using a single prompt for each client that is aggregated and updated globally on the server may lead to overfitting on shared features, thereby failing to capture the unique characteristics of individual client datasets. To mitigate this issue, we propose a dual prompt mechanism, comprising a \textit{Shared Prompt} and a \textit{Private Prompt}, that balances global consistency with local adaptability.

\textit{Shared Prompt Learning.} The {Shared Prompt} \( s_g \) is designed to capture common features across all clients and is updated globally. During federated learning, each client \( \mathcal{C}_i \) locally updates its shared prompt \( s_{g,i} \) and transmits it to the central server. The server aggregates these prompts from all clients to produce an updated global shared prompt, which is distributed back to the clients. The aggregation of the shared prompts can be formulated as:
\begin{equation}
s_g^{(t+1)} = \frac{1}{N} \sum_{i=1}^{N}  w_i s_{g,i}^{(t)}
\end{equation}
where \( s_g^{(t+1)} \) is the global shared prompt at round \( t+1 \), \( N \) is the number of participating clients, and \( s_{g,i}^{(t)} \) represents the shared prompt for client \( \mathcal{C}_i \) at round \( t \). The primary objective of the shared prompt is to capture generalized knowledge across clients, enabling the model to better generalize to new data across different client distributions. Through this mechanism, the model learns global patterns that are critical for handling diverse datasets across all clients.

\textit{Private Prompt Learning.} In contrast, the {Private Prompt} \( s_{p,i} \) is specific to each client \( \mathcal{C}_i \) and captures the unique characteristics of the local data. Given that the data distribution varies between clients, the private prompt enables each client to fine-tune the model to better align with its local dataset. Unlike the shared prompt, the private prompt is optimized locally and is not shared with the central server, thereby preserving data privacy while enabling personalized model adjustments.

The final prediction for an input \( I_j \) on client \( \mathcal{C}_i \) is generated by combining the shared prompt \( s_g \) and the private prompt \( s_{p,i} \). This approach leverages global knowledge from \( s_g \) while allowing for client-specific adaptation via \( s_{p,i} \). The prediction for a given input \( I_j \) can be formulated as:
\begin{equation}
\hat{y}_j = f([s_g, s_{p,i}]; I_j)
\end{equation}
where \( \hat{y}_j \) is the predicted label for the input image \( I_j \), and \( f([s_g, s_{p,i}]; I_j) \) represents the model's decision function based on the combination of the shared and private prompts. The private prompt \( s_{p,i} \) is trained using the local dataset \( \mathcal{D}_i \), with the following loss function \( \mathcal{L}_{p,i} \):
\begin{equation}
\mathcal{L}_{p,i} = \frac{1}{|\mathcal{D}_i|} \sum_{(I_j, y_j) \in \mathcal{D}_i} \ell(\hat{y}_j, y_j)
\end{equation}
where \( \ell(\hat{y}_j, y_j) \) is a loss function (e.g., cross-entropy) between the predicted label \( \hat{y}_j \) and the ground-truth label \( y_j \). This loss function enables each client to tailor its model to its local data while preserving the global knowledge derived from the shared prompt.

\textit{Federated Prompt Updates.} During each round of federated learning, the shared prompt \( s_g \) is updated across all clients, while each client independently updates its private prompt \( s_{p,i} \) based on its local data. The shared prompt ensures model consistency across the federation, while the private prompt provides the flexibility to adapt to individual client needs. This iterative process allows the model to continuously learn from both global and local data, achieving a balance between global consistency and local adaptation. Over multiple rounds, this framework fosters a model that generalizes well across heterogeneous client environments while still addressing specific local needs.

\subsection{Dual Prompt Alignment Constraint}
As described in the last section, we introduce the shared prompt and private prompt mechanisms to address the challenge of balancing global consistency and local adaptability in federated learning. The shared prompt is designed to capture global features across clients, ensuring that the model learns shared information from all client data. In contrast, the private prompt is tailored to each client’s specific data, allowing the model to adapt to diverse local data distributions. To ensure that the private prompt and the shared prompt remain semantically aligned during the learning process, we propose the \textit{Dual Prompt Alignment Constraint}. This constraint acts as a loss function that aligns the features learned by the shared and private prompts, maintaining semantic coherence while balancing global and local learning objectives.

In each client, the private prompt feature is denoted as \( E_{Tp,i} \), while the shared prompt feature is represented as \( E_{Ts} \). To ensure that each client’s private prompt feature \( E_{Tp,i} \) aligns with its corresponding shared prompt feature \( E_{Ts} \), we define the following alignment constraint loss function. Specifically, \( \varphi \) and \( \psi \) are embedding functions that map the private prompt feature \( E_{Tp,i} \) and the shared prompt feature \( E_{Ts} \), respectively, into a common feature space where alignment is enforced. This loss encourages each client’s private prompt feature to be closer to its corresponding shared prompt feature than to the shared prompts of other clients:
\begin{equation}
\begin{aligned}
L_{PAC}^{i} = \log\Bigg( 1 + \sum_{j \neq i} \exp\bigg[ s \cdot \varphi(E_{Tp,i})^\top \psi(E_{Ts,j}) \\
- s \cdot \varphi(E_{Tp,i})^\top \psi(E_{Ts}) \bigg] \Bigg)
\end{aligned}
\end{equation}
where \( E_{Tp,i} \) represents the private prompt features of the \(i\)-th client, \( E_{Ts} \) represents the global shared prompt features, and \( E_{Ts,j} \) represents the shared prompt features of other clients. The embedding functions \( \varphi \) and \( \psi \) ensure that both private and shared prompt features are mapped into a shared feature space for effective comparison. The parameter \( s \) serves as a scaling factor to modulate the alignment. By minimizing this loss, we ensure that each client’s private prompt feature is aligned with its corresponding global shared prompt, preventing significant deviations. To generalize this across the federated learning framework, we aggregate the alignment losses from all clients into a global optimization objective. The overall dual prompt alignment constraint is defined as:
\begin{equation}\small
L_{PAC} = \frac{1}{N} \sum_{i=1}^{N} \log\left( \frac{ \exp\left( s \cdot \varphi(E_{Tp,i})^\top \psi(E_{Ts}) \right) }{ \sum_{j \neq i} \exp\left( s \cdot \varphi(E_{Tp,i})^\top \psi(E_{Ts,j}) \right) } \right)
\end{equation}
where \( N \) is the total number of clients. By minimizing this loss function, the federated learning system updates both the shared prompt and the private prompt during each communication round, ensuring semantic alignment between global and local prompts. This optimization allows each client’s model to effectively incorporate shared global knowledge while remaining flexible to adapt to the unique characteristics of local data distributions.

The dual prompt alignment constraint provides an efficient mechanism for maintaining semantic relevance between the shared and private prompts. It ensures that private prompts can adapt to local data without deviating from the semantic structure of global features. This approach employs a metric-learning-based strategy to align features, achieving semantic coherence while minimizing additional computational overhead. Furthermore, it enhances the robustness of federated learning by improving semantic alignment across heterogeneous data distributions, ultimately promoting better generalization and performance across clients.

\subsection{Cross-Modal Feature Alignment Constraint}
To further align the text features \(E_T\), including both the shared and private features obtained through the text encoder, with the image features \(E_I\) in our modified CLIP model, we employ \textit{Cross-Modal Feature Alignment Constraint} to handle the alignment between the two modalities. This constraint addresses the alignment challenges between the two modalities by leveraging Optimal Transport algorithm~\cite{chizat2018scaling}, enabling FedRSCLIP to effectively align cross-modal features through the cooperative use of global and local prompts. Specifically, we define the cost matrix \(C\) as the cosine distance between the image and text features, which is formulated as:
\begin{equation}
C = 1 - \cos(E_I, E_T)
\end{equation}
where \(C\) is a matrix based on cosine distance, representing the discrepancy between the image features \(E_I\) and the text features \(E_T\). Cosine distance is used to measure the difference between the two feature vectors, where a smaller value indicates greater similarity. Inspired by~\cite{fedotp}, we introduced an entropic regularization term into the original optimal transport objective to ensure that the prompts focus on relevant regions of the image, avoiding capturing irrelevant information. The final optimization objective is given by:
\begin{equation}
d_{C,k}(\alpha, \beta) = \min_{T \in U(\alpha, \beta)} \langle C, T \rangle + \lambda \langle T, \log T \rangle
\end{equation}
where \( U(\alpha, \beta) \) is the constraint set of the transport plan, defined as:
\begin{equation}
U(\alpha, \beta) = \left\{ T \in \mathbb{R}_+^{V \times 2} \mid T \mathbf{1}_2 \leq \alpha, T^\top \mathbf{1}_V = \beta \right\}
\end{equation}
Here, \(T\) represents the transport plan matrix, while \( \alpha \in \mathbb{R}^V \) and \( \beta \in \mathbb{R}^2 \) are the marginal distributions of the image and text features. By adjusting the size of these marginal distributions, we can flexibly control the mapping between the prompts and the image feature map. Referring to~\cite{benamou2015iterative}, we further reformulate the above objective as a Kullback-Leibler (KL) projection. To accelerate the optimization of the transport plan, we adopt the fast implementation of the Dykstra algorithm~\cite{dykstra1983algorithm}. Through the Dykstra algorithm, we efficiently scale the KL projection to rapidly solve the transport plan. After initializing \( Q = \exp(-C/\lambda) \) and \( v(0) = \mathbf{1}_2 \), the optimal solution is obtained through the following iterative update:
\begin{equation}
T^* = \text{diag}(u(\tilde{t})) Q \text{diag}(v(\tilde{t}))
\end{equation}
where \(u(\tilde{t})\) and \(v(\tilde{t})\) are updated vectors at each iteration, with the initial value \(v(0)\) set as a vector of ones. After a few iterations, we quickly converge to the optimal transport plan \(T^*\). Once the optimal transport plan \(T^*\) is obtained, we compute the Wasserstein distance \(d_{C,k}\), and the prediction probability \(p\) is rewritten from Equation (5) as follows:
\begin{equation}
p(y = k \mid I_j) = \frac{\exp((1 - d_{C,k}) / \tau)}{\sum_{c=1}^{K} \exp((1 - d_{C,c}) / \tau)}
\end{equation}
After obtaining the prediction probability \(p\), we fix the transport plan \(T^*\) and simultaneously optimize the learnable vectors in the shared and private prompts to ensure precise alignment between the visual and textual features, thereby improving both global generalization and local adaptability.

\section{Experiments}

\subsection{Experimental Setting}

To validate the effectiveness and advancements of our proposed method in remote sensing image classification, we benchmarked against several state-of-the-art federated learning algorithms. Specifically, we selected the original FedAvg\cite{fedavg}, which transmits the complete set of model parameters, as well as FedAvg, FedProx\cite{fedprox}, and FedOTP\cite{fedotp} with Prompt Learning. For the original FedAvg, which does not utilize a Vision-Language Model (VLM) structure and operates as a standard image classification network, we followed the settings outlined in\cite{ben2024federated}. These include Vision Transformer-based models (ViT-Tiny and ViT-Base) and CNN-based models (EfficientNet-B1 and EfficientNet-B3). In contrast, FedAvg, FedProx, FedOTP, and our model with Prompt Learning are all based on the ViT-Base backbone CLIP framework.

In our experiments, all input images from the three datasets were resized to 224 × 224 pixels and divided into 14 × 14 patches, each with a feature dimension of 768. For optimization, we utilized the Stochastic Gradient Descent (SGD) optimizer with an initial learning rate of 0.001. The batch size was set to 32 for training and 100 for testing. All experiments were implemented using the PyTorch framework and conducted on NVIDIA 3090 GPUs to ensure computational consistency.

\begin{figure*}[t]
  \centering
    \includegraphics[width=1.\textwidth]{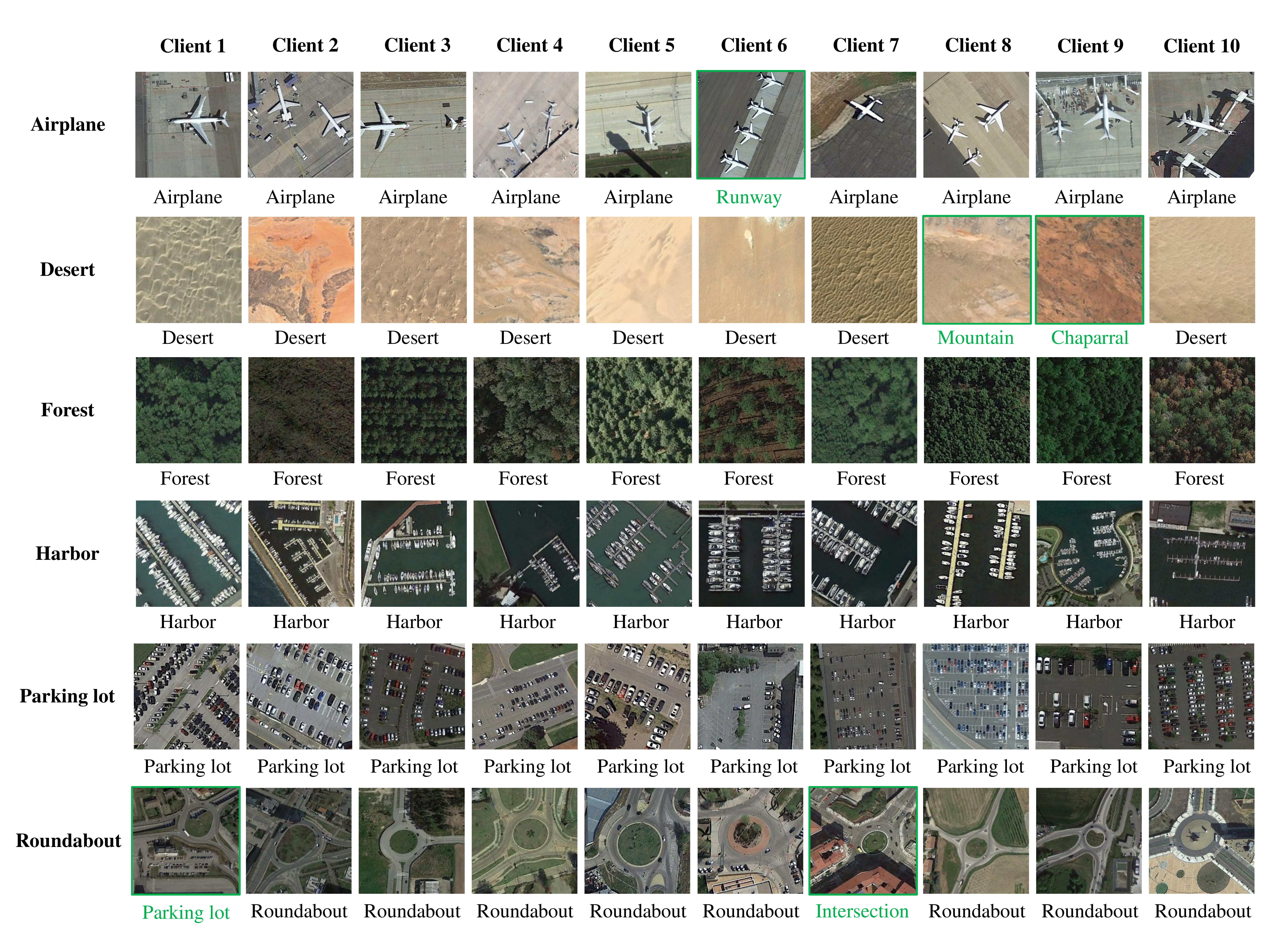}
  \caption{Classification results of FedRSClip across 10 clients on the Fed-Optimal dataset. Six representative classes are randomly selected for visualization: Airplane, Desert, Forest, Harbor, Parking Lot, and Roundabout. Each column corresponds to a different client, and the rows represent the predicted class labels. Correct classifications are shown in their respective rows, while misclassified samples are highlighted with green borders. 
  }
  \label{fig:res_opti}
\end{figure*}

\begin{table*}[t]
\centering
\caption{Performance comparison of various models on the Fed-Optimal dataset across centralized and federated learning configurations with different numbers of clients. The last column indicates the parameter size transmitted between the server and clients for each model.}
\label{tab:res_opti}
\resizebox{\textwidth}{!}{
\begin{tabular}{lcccccccc}
\toprule
\multirow{2}{*}{\textbf{Model Type}} & \multicolumn{7}{c}{\textbf{Number of Clients}} & \multirow{2}{*}{\textbf{Transmitted Parameter}} \\
& \textbf{Centralized} & \textbf{2} & \textbf{5} & \textbf{10} & \textbf{15} & \textbf{20} & \textbf{40} & \\ 
\midrule
FedAvg~\cite{fedavg} (ViT-Tiny) & 87.05 & 84.67 & 87.01 & 85.23 & 82.41 & 79.96 & 70.60 & 5.7 M \\    
FedAvg~\cite{fedavg} (ViT-Base) & 89.46 & 82.09 & 88.04 & 90.60 & 92.69 & {92.22} & 91.94 & 86.6 M \\    
FedAvg~\cite{fedavg} (EfficientNet-B1) & 87.55 & 88.95 & 86.86 & 83.76 & 77.72 & 69.57 & 34.69 & 7.8 M\\    
FedAvg~\cite{fedavg} (EfficientNet-B3) & 87.05 & 87.96 & 86.82 & 83.87 & 77.40 & 68.19 & 32.43 & 12.0 M \\ 
FedAvg-PL~\cite{fedavg} & 91.61 & 91.53 & 91.33 & 89.99 & 89.70 & 89.79 & 91.13 & 2 K \\
FedProx~\cite{fedprox} & 91.94 & 92.63 & 90.67& 92.24 & 89.80 & 91.18 & 91.89 & 2 K \\
FedOTP~\cite{fedotp} & 91.78 & 92.84 & 92.80 & 93.00 & 92.45 & 90.80 & 93.46 & 2 K \\
FedRSClip (Ours) & \textbf{92.90} & \textbf{93.38} & \textbf{94.52} & \textbf{94.47} & \textbf{94.05} & \textbf{93.83} & \textbf{94.46} & 2 K  \\
\bottomrule
\end{tabular}}
\end{table*}

\subsection{Results}

\subsubsection{Results on Fed-Optimal Dataset}

Table~\ref{tab:res_opti} provides a comprehensive performance comparison of various models on the Fed-Optimal dataset across centralized and federated learning configurations, along with the parameter sizes transmitted between the server and clients. Among the baseline methods, the model parameter transfer-based federated learning approach, FedAvg, and its variants demonstrate a noticeable decline in performance as the number of clients increases, highlighting their sensitivity to client heterogeneity and data fragmentation. For instance, FedAvg (EfficientNet-B1) achieves 87.55\%in centralized settings but drops sharply to 34.69\%with 40 clients, while requiring the transmission of 7.8M parameters between the server and clients. In contrast, FedAvg-PL, which integrates prompt learning, significantly outperforms traditional FedAvg variants by achieving 91.61\%in centralized setups and maintaining superior performance across federated configurations, all while reducing the transmitted parameter size to 2,048 parameters via the prompt learning mechanism. This result underscores the effectiveness of prompt learning in mitigating challenges inherent in federated optimization. Similarly, FedProx and FedOTP further improve performance compared to earlier FedAvg-based methods, achieving better stability and accuracy under increasing numbers of clients. Compared to the aforementioned methods, our proposed FedRSClip achieves the best overall performance across all scenarios, demonstrating its significant advantages in both centralized and federated learning configurations. Specifically, FedRSClip consistently maintains exceptional accuracy as the number of clients increases, with only minimal performance degradation even in highly distributed settings. For instance, FedRSClip achieves 93.38\% accuracy with 2 clients, 94.52\% with 5 clients, and 94.46\% with 40 clients, while transmitting only 2,048 parameters. These results highlight the remarkable scalability, robustness, and efficiency of FedRSClip, showcasing its ability to effectively mitigate the challenges posed by client heterogeneity and data imbalance, which are common in real-world federated learning scenarios. Additionally, the compactness of its transmitted parameter size further establishes FedRSClip as a practical and scalable solution for large-scale federated learning applications.

To intuitively showcase the classification results, we visualize the predictions of FedRSClip across 10 clients on the Fed-Optimal dataset, as shown in Fig.~\ref{fig:res_opti}. Six representative classes, including Airplane, Desert, Forest, Harbor, Parking Lot, and Roundabout, are randomly selected for this visualization. Each column represents a specific client, while each row corresponds to the predicted class labels. Correct classifications are shown in their respective rows, and misclassified samples are highlighted with green borders. Through a careful analysis of the misclassified samples, we observe that these errors can be categorized into two distinct cases. The first case arises from the high similarity of features between different scenes, leading to classification errors. For example, in the Desert class, Client 8 misclassifies the image as "Mountain," and Client 9 as "Chaparral." These errors are primarily due to the inherent visual similarity in texture and color patterns between these scenes, which poses challenges for fine-grained differentiation. Similarly, in the Roundabout class, Client 7 predicts "Intersection," likely due to the structural resemblance between roundabouts and intersections in urban planning. The second case involves misclassifications that, although not matching the ground truth label, are semantically consistent with the image content. For instance, in the Airplane class, Client 6 predicts "Runway," and the image indeed contains a visible runway, demonstrating that the model captures underlying semantic information. Similarly, in the Roundabout class, Client 1 predicts "Parking Lot," which is partially correct as the image includes parking spaces near the roundabout.

Overall, these results highlight the robustness and adaptability of FedRSClip in federated learning scenarios. Despite client heterogeneity and data imbalance, the model maintains high classification accuracy across multiple clients and demonstrates its ability to handle complex and ambiguous scenes. These advantages further validate FedRSClip's exceptional performance and potential for distributed learning applications.

\subsubsection{Results on Fed-UCMerced Dataset}
We further evaluated the performance of our proposed method on the Fed-UCMerced dataset, as shown in Table~\ref{tab:res_ucm}, which provides a performance comparison of our method and baseline approaches across centralized and federated learning configurations. Considering the generally poor performance of traditional parameter transfer-based federated learning approachs observed in Table~\ref{tab:res_opti}, we only include the higher-performing baselines, including FedAvg-PL, FedProx, and FedOTP, for this comparison. As shown in Table~\ref{tab:res_ucm}, FedRSClip consistently achieves the highest accuracy across all scenarios, demonstrating state-of-the-art performance in both centralized and federated settings. Specifically, FedRSClip achieves an accuracy of 96.38\% in centralized training and continues to outperform other methods as the number of clients increases, maintaining an accuracy of 96.86\% with 2 clients, 96.38\% with 5 clients, and 95.95\% with 40 clients. This consistent performance highlights its robustness and scalability under diverse client distributions. Furthermore, FedRSClip exhibits superior scalability, with only a marginal drop in accuracy (0.43\%) from centralized to the most distributed configuration (40 clients). These results underscore the resilience of FedRSClip in federated learning scenarios, effectively addressing challenges such as client heterogeneity and data imbalance. Overall, the results establish FedRSClip as a robust and scalable solution for large-scale federated learning applications, providing consistent high performance even in challenging distributed environments.

\begin{table}[h]
\centering
\caption{Performance comparison of various models on the Fed-UCMerced dataset across centralized and federated learning configurations with different numbers of clients.}
\label{tab:res_ucm}
\resizebox{0.5\textwidth}{!}{
\begin{tabular}{lccccccc}
\toprule
\multirow{2}{*}{\textbf{Model Type}} & \multicolumn{7}{c}{\textbf{Number of Clients}} \\
& \textbf{Centralized} & \textbf{2} & \textbf{5} & \textbf{10} & \textbf{15} & \textbf{20} & \textbf{40} \\ 
\midrule
FedAvg-PL~\cite{fedavg} & 95.43 & 95.51 & 95.10 & 95.15 & 93.84 & 94.87  & 94.84 \\
FedProx~\cite{fedprox} & 95.95 & 95.53 & 95.95 & 95.36 & 95.12 & 95.11 & 95.21 \\
FedOTP~\cite{fedotp} & 95.81 & 96.08 & 96.06 & 96.49 &95.70 & 95.27& 95.37 \\
FedRSClip & \textbf{96.38} & \textbf{96.86} & \textbf{96.38} & \textbf{96.88} & \textbf{95.84} & \textbf{96.11} & \textbf{95.95} \\
\bottomrule
\end{tabular}}
\end{table}

\subsubsection{Results on Fed-NWPU Dataset}

Finally, we evaluated the performance of our method on the Fed-NWPU dataset, as shown in Table~\ref{tab:res_nwpu}. It is important to note that for this dataset, we only utilized 20\% of the training data, significantly increasing the difficulty of the task compared to the previous two datasets. The reduced training data not only intensifies the heterogeneity and imbalance in data distribution but also places greater demands on the model's learning capability. Consistent with previous experiments, we selected high-performing baselines, including FedAvg-PL, FedProx, and FedOTP, for comparison. The results demonstrate that despite the increased task difficulty, FedRSClip achieves the highest accuracy across all configurations, further validating its robustness and scalability. In centralized training, FedRSClip achieves the best accuracy of 90.03\%, slightly outperforming FedAvg-PL (89.91\%), FedProx (89.94\%), and FedOTP (88.41\%). In federated learning scenarios, FedRSClip maintains its leading position and demonstrates stable performance as the number of clients increases. For instance, FedRSClip achieves an accuracy of 91.55\% with 2 clients, 92.08\% with 5 clients, and 90.59\% with 40 clients. These results indicate that even under the dual challenges of reduced training data and diverse data distributions, FedRSClip effectively mitigates performance degradation. In contrast, other baseline methods show more significant performance declines. For example, FedOTP's accuracy drops from 91.49\% with 2 clients to 90.27\% with 40 clients, whereas FedRSClip experiences a smaller decline, highlighting its superior scalability in federated learning scenarios. Moreover, FedRSClip consistently maintains high performance across different configurations, demonstrating its adaptability to heterogeneous data and client environments as well as its resilience to challenges such as data imbalance. These results further confirm FedRSClip's outstanding performance on the NWPU dataset, establishing it as a robust and scalable solution for both centralized and federated learning setups.

\begin{table}[h]
\centering
\caption{Performance comparison of various models on the Fed-NWPU dataset across centralized and federated learning configurations with different numbers of clients.}
\label{tab:res_nwpu}
\resizebox{0.5\textwidth}{!}{
\begin{tabular}{lccccccc}
\toprule
\multirow{2}{*}{\textbf{Model Type}} & \multicolumn{7}{c}{\textbf{Number of Clients}} \\
& \textbf{Centralized} & \textbf{2} & \textbf{5} & \textbf{10} & \textbf{15} & \textbf{20} & \textbf{40} \\ 
\midrule
FedAvg-PL~\cite{fedavg} & 89.91 & 88.00 & 88.42 & 88.78 & 88.16 & 88.46 & 88.76 \\
FedProx~\cite{fedprox} & 89.94 & 91.47 & 91.60 & 90.77 & 90.50 & 90.65 & 90.16  \\
FedOTP~\cite{fedotp} & 88.41 &91.49 &91.74 & 90.81 & 90.57 & 90.85 & 90.27 \\
FedRSClip & \textbf{90.03} & \textbf{91.55} & \textbf{92.08} &  \textbf{90.97} & \textbf{90.99} & \textbf{91.06} & \textbf{90.59}\\
\bottomrule
\end{tabular}}
\end{table}

\subsection{Ablation Study}

\subsubsection{Effectiveness of Dual Prompt Mechanism}

To demonstrate the effectiveness of the proposed dual-prompt mechanism, we conducted comparative experiments evaluating FedRSClip with the Standard Prompt Mechanism (SPM) and Dual Prompt Mechanism (DPM) across centralized and federated learning configurations. As shown in Table~\ref{tab:res_ab_dpm}, our model with DPM significantly outperforms the SPM variant across all configurations, demonstrating the superiority of the proposed mechanism. In centralized settings, the DPM-enabled model achieves an accuracy of 92.90\%, surpassing the SPM-enabled variant by 1.08 percentage points. This performance gap becomes even more pronounced in federated settings. For example, with 10 clients, the DPM-enabled model achieves an impressive accuracy of 94.47\%, compared to 90.13\% for the SPM-enabled model, a notable improvement of 4.34 percentage points. The superiority of the dual prompt mechanism stems from its design, which integrates shared prompts to capture global features and private prompts to adapt to local data distributions. This dual structure allows the model to effectively balance global consistency with local adaptability, addressing the complexities of federated learning environments. In comparison, the standard prompt mechanism limits its flexibility, resulting in suboptimal performance, particularly in highly distributed scenarios. Overall, the results validate the effectiveness of the proposed DPM in mitigating the challenges of federated learning, establishing it as a crucial innovation in FedRSClip. By leveraging DPM, FedRSClip achieves state-of-the-art performance across both centralized and federated configurations, demonstrating its robustness, scalability, and adaptability to diverse data distributions.

\begin{table}[h]
\centering
\caption{Performance comparison of FedRSClip with Standard Prompt Mechanism (SPM) and Dual Prompt Mechanism (DPM) across centralized and federated learning configurations with different numbers of clients.}
\label{tab:res_ab_dpm}
\resizebox{0.5\textwidth}{!}{
\begin{tabular}{lccccccc}
\toprule
\multirow{2}{*}{\textbf{Model Type}} & \multicolumn{7}{c}{\textbf{Number of Clients}} \\
& \textbf{Centralized} & \textbf{2} & \textbf{5} & \textbf{10} & \textbf{15} & \textbf{20} & \textbf{40} \\ 
\midrule
FedRSClip w/ SPM  & 91.82 & 92.54 & 91.61 & 90.13 & 90.09 & 87.67 & 90.65 \\    
FedRSClip w/ DPM & \textbf{92.90} & \textbf{93.38} & \textbf{94.52} & \textbf{94.47} & \textbf{94.05} & \textbf{93.83} & \textbf{94.46} \\  
\bottomrule
\end{tabular}
}
\end{table}

\subsubsection{Effectiveness of Dual Prompt Alignment Constraint}

To assess the impact of the Dual Prompt Alignment Constraint (DPAC), we performed an ablation study comparing the performance of FedRSClip with and without DPAC across centralized and federated learning configurations, as detailed in Table VI. The results demonstrate that the inclusion of DPAC leads to noticeable improvements in accuracy across all configurations. In the centralized scenario, FedRSClip equipped with DPAC achieves an accuracy of 92.90\%, surpassing the 91.93\% achieved without DPAC. The advantages of DPAC become even more prominent in federated settings. For example, when using 10 clients, the model incorporating DPAC attains an accuracy of 94.47\%, outperforming its counterpart without DPAC, which reaches 93.14\%. Similarly, with 40 clients, FedRSClip with DPAC achieves 94.46\%, representing a notable improvement over the 93.64\% recorded without DPAC. These findings underline the importance of DPAC in improving the alignment of private and shared prompts, which facilitates a better balance between global consistency and local adaptability. By addressing challenges such as non-iid data and ensuring robust generalization across diverse clients, DPAC serves as a crucial enhancement for FedRSClip, significantly boosting its scalability and performance in federated learning environments.

\begin{table}[h]
\centering
\caption{Performance comparison of FedRSClip with and without Dual Prompt Alignment Constraint (DPAC) across centralized and federated learning configurations with different numbers of clients.}
\resizebox{0.5\textwidth}{!}{
\begin{tabular}{lccccccc}
\toprule
\multirow{2}{*}{\textbf{Model Type}} & \multicolumn{7}{c}{\textbf{Number of Clients}} \\
& \textbf{Centralized} & \textbf{2} & \textbf{5} & \textbf{10} & \textbf{15} & \textbf{20} & \textbf{40} \\ 
\midrule
FedRSClip w/o DPAC  & 91.93 & 93.15 & 93.24 & 93.14 & 93.88 & 92.87 & 93.64 \\    
FedRSClip w/ DPAC & \textbf{92.90} & \textbf{93.38} & \textbf{94.52} & \textbf{94.47} & \textbf{94.05} & \textbf{93.83} & \textbf{94.46} \\  
\bottomrule
\end{tabular}
}
\end{table}

\subsubsection{Effectiveness of Cross-Modal Feature Alignment Constraint}
To validate the effectiveness of the proposed Cross-Modal Feature Alignment Constraint (CMFAC), we conducted an ablation study by comparing FedRSClip with and without CMFAC across centralized and federated learning configurations, as shown in Table V. The results demonstrate that incorporating CMFAC consistently improves performance in both centralized and federated scenarios. For example, in the centralized setting, FedRSClip with CMFAC achieves an accuracy of 92.90\%, compared to 92.36\% without CMFAC, reflecting a notable improvement. This trend is more evident in federated configurations, where the challenges of client heterogeneity and data imbalance are more pronounced. With 10 clients, the model with CMFAC achieves 94.47\%, outperforming the version without CMFAC, which achieves 92.74\%. Similarly, with 40 clients, the performance improvement is substantial, with the model achieving 94.46\% with CMFAC compared to 93.08\% without CMFAC. These results highlight the critical role of CMFAC in aligning multimodal features effectively, enabling better generalization across diverse client distributions and mitigating performance degradation caused by non-iid data. The consistent improvements across all configurations validate CMFAC as a key component in enhancing the robustness and scalability of FedRSClip in federated learning environments.

\begin{table}[h]
\centering
\caption{Performance comparison of FedRSClip with and without Cross-Modal Feature Alignment Constraint (CMFAC) across centralized and federated learning configurations with different numbers of clients.}
\resizebox{0.5\textwidth}{!}{
\begin{tabular}{lccccccc}
\toprule
\multirow{2}{*}{\textbf{Model Type}} & \multicolumn{7}{c}{\textbf{Number of Clients}} \\
& \textbf{Centralized} & \textbf{2} & \textbf{5} & \textbf{10} & \textbf{15} & \textbf{20} & \textbf{40} \\ 
\midrule
FedRSClip w/o CMFAC  & 92.36 & 93.01 & 93.30 & 92.74 & 93.59 & 92.57 & 93.08 \\    
FedRSClip w/ CMFAC & \textbf{92.90} & \textbf{93.38} & \textbf{94.52} & \textbf{94.47} & \textbf{94.05} & \textbf{93.83} & \textbf{94.46} \\  
\bottomrule
\end{tabular}
}
\end{table}

\section{Conclusion}
In this paper, we introduced FedRSCLIP, a federated learning framework for remote sensing image classification built upon the ViT-Base backbone CLIP framework. To the best of our knowledge, this is the first framework to seamlessly integrate VLMs within federated learning for remote sensing tasks. FedRSCLIP effectively addresses challenges associated with data heterogeneity and the inefficiencies of transmitting large-scale language-vision models in federated environments. By incorporating Prompt Learning, our framework reduces communication overhead by optimizing a small set of tunable parameters while retaining adaptability to local client data. The use of Shared Prompts for global knowledge sharing and Private Prompts for client-specific adaptation ensures a balance between global consistency and local flexibility. To further enhance alignment and adaptability, we introduced the Prompt Alignment Constraint Loss, which maintains semantic coherence between shared and private prompts, and the Cross-Modal Feature Alignment Constraint, which ensures effective alignment between textual and image features. Extensive experiments conducted on our constructed FedRSIC dataset demonstrate that FedRSCLIP achieves state-of-the-art performance. Our model improves classification accuracy and communication efficiency in remote sensing image classification tasks, showcasing its robustness across diverse federated learning configurations. This work provides a solid foundation for advancing the integration of VLMs into federated learning frameworks, particularly in scenarios involving large-scale, heterogeneous data. Future research can build on this foundation to explore further enhancements in cross-modal alignment and adaptivity in federated environments.

{\small
\bibliographystyle{IEEEtran}
\bibliography{egbib}

\begin{thebibliography}{10}
\providecommand{\url}[1]{#1}
\csname url@samestyle\endcsname
\providecommand{\newblock}{\relax}
\providecommand{\bibinfo}[2]{#2}
\providecommand{\BIBentrySTDinterwordspacing}{\spaceskip=0pt\relax}
\providecommand{\BIBentryALTinterwordstretchfactor}{4}
\providecommand{\BIBentryALTinterwordspacing}{\spaceskip=\fontdimen2\font plus
\BIBentryALTinterwordstretchfactor\fontdimen3\font minus \fontdimen4\font\relax}
\providecommand{\BIBforeignlanguage}[2]{{%
\expandafter\ifx\csname l@#1\endcsname\relax
\typeout{** WARNING: IEEEtran.bst: No hyphenation pattern has been}%
\typeout{** loaded for the language `#1'. Using the pattern for}%
\typeout{** the default language instead.}%
\else
\language=\csname l@#1\endcsname
\fi
#2}}
\providecommand{\BIBdecl}{\relax}
\BIBdecl

\bibitem{tian2023shape}
F.~Tian, B.~Wu, H.~Zeng, M.~Zhang, Y.~Hu, Y.~Xie, C.~Wen, Z.~Wang, X.~Qin, W.~Han \emph{et~al.}, ``A shape-attention pivot-net for identifying central pivot irrigation systems from satellite images using a cloud computing platform: an application in the contiguous us,'' \emph{GIScience \& Remote Sensing}, vol.~60, no.~1, p. 2165256, 2023.

\bibitem{han2024autoencoding}
W.~Han, C.~Wen, L.~Chok, Y.~L. Tan, S.~L. Chan, H.~Zhao, and C.~Feng, ``Autoencoding tree for city generation and applications,'' \emph{ISPRS Journal of Photogrammetry and Remote Sensing}, vol. 208, pp. 176--189, 2024.

\bibitem{rwanga2017accuracy}
S.~S. Rwanga, J.~M. Ndambuki \emph{et~al.}, ``Accuracy assessment of land use/land cover classification using remote sensing and gis,'' \emph{International Journal of Geosciences}, vol.~8, no.~04, p. 611, 2017.

\bibitem{wen2019novel}
C.~Wen, S.~Liu, X.~Yao, L.~Peng, X.~Li, Y.~Hu, and T.~Chi, ``A novel spatiotemporal convolutional long short-term neural network for air pollution prediction,'' \emph{Science of the total environment}, vol. 654, pp. 1091--1099, 2019.

\bibitem{advances}
P.~Kairouz, H.~B. McMahan, B.~Avent, A.~Bellet, M.~Bennis, A.~N. Bhagoji, K.~Bonawitz, Z.~Charles, G.~Cormode, R.~Cummings \emph{et~al.}, ``Advances and open problems in federated learning,'' \emph{Foundations and trends{\textregistered} in machine learning}, 2021.

\bibitem{fedavg}
B.~McMahan, E.~Moore, D.~Ramage, S.~Hampson, and B.~A. y~Arcas, ``Communication-efficient learning of deep networks from decentralized data,'' in \emph{Artificial intelligence and statistics}.\hskip 1em plus 0.5em minus 0.4em\relax PMLR, 2017, pp. 1273--1282.

\bibitem{fedprox}
T.~Li, A.~K. Sahu, M.~Zaheer, M.~Sanjabi, A.~Talwalkar, and V.~Smith, ``Federated optimization in heterogeneous networks,'' \emph{Proceedings of Machine learning and systems}, 2020.

\bibitem{fedalign}
M.~Mendieta, T.~Yang, P.~Wang, M.~Lee, Z.~Ding, and C.~Chen, ``Local learning matters: Rethinking data heterogeneity in federated learning,'' in \emph{Proceedings of the IEEE Conference on Computer Vision and Pattern Recognition (CVPR)}, 2022, pp. 8397--8406.

\bibitem{zhu2023minigpt}
D.~Zhu, J.~Chen, X.~Shen, X.~Li, and M.~Elhoseiny, ``Minigpt-4: Enhancing vision-language understanding with advanced large language models,'' \emph{arXiv preprint arXiv:2304.10592}, 2023.

\bibitem{dai2023instructblip}
W.~Dai, J.~Li, D.~Li, A.~M.~H. Tiong, J.~Zhao, W.~Wang, B.~Li, P.~Fung, and S.~Hoi, ``Instructblip: Towards general-purpose vision-language models with instruction tuning,'' \emph{arXiv preprint arXiv:2305.06500}, 2023.

\bibitem{liu2024visual}
H.~Liu, C.~Li, Q.~Wu, and Y.~J. Lee, ``Visual instruction tuning,'' \emph{Advances in neural information processing systems}, vol.~36, 2024.

\bibitem{clip}
A.~Radford, J.~W. Kim, C.~Hallacy, A.~Ramesh, G.~Goh, S.~Agarwal, G.~Sastry, A.~Askell, P.~Mishkin, J.~Clark \emph{et~al.}, ``Learning transferable visual models from natural language supervision,'' in \emph{International conference on machine learning}.\hskip 1em plus 0.5em minus 0.4em\relax PMLR, 2021, pp. 8748--8763.

\bibitem{hu2023rsgpt}
Y.~Hu, J.~Yuan, C.~Wen, X.~Lu, and X.~Li, ``Rsgpt: A remote sensing vision language model and benchmark,'' \emph{arXiv preprint arXiv:2307.15266}, 2023.

\bibitem{bazi2024rs}
Y.~Bazi, L.~Bashmal, M.~M. Al~Rahhal, R.~Ricci, and F.~Melgani, ``Rs-llava: A large vision-language model for joint captioning and question answering in remote sensing imagery,'' \emph{Remote Sensing}, vol.~16, no.~9, p. 1477, 2024.

\bibitem{pang2024h2rsvlm}
C.~Pang, J.~Wu, J.~Li, Y.~Liu, J.~Sun, W.~Li, X.~Weng, S.~Wang, L.~Feng, G.-S. Xia, and C.~He, ``H2rsvlm: Towards helpful and honest remote sensing large vision language model,'' 2024.

\bibitem{kuckreja2023geochat}
K.~Kuckreja, M.~S. Danish, M.~Naseer, A.~Das, S.~Khan, and F.~S. Khan, ``Geochat: Grounded large vision-language model for remote sensing,'' 2023.

\bibitem{lin2024rs}
H.~Lin, D.~Hong, S.~Ge, C.~Luo, K.~Jiang, H.~Jin, and C.~Wen, ``Rs-moe: Mixture of experts for remote sensing image captioning and visual question answering,'' \emph{arXiv preprint arXiv:2411.01595}, 2024.

\bibitem{noniiddata}
Q.~Li, Y.~Diao, Q.~Chen, and B.~He, ``Federated learning on non-iid data silos: An experimental study,'' in \emph{2022 IEEE 38th international conference on data engineering (ICDE)}.\hskip 1em plus 0.5em minus 0.4em\relax IEEE, 2022, pp. 965--978.

\bibitem{timeevolving1}
Y.~Guo, T.~Lin, and X.~Tang, ``Towards federated learning on time-evolving heterogeneous data,'' \emph{arXiv preprint arXiv:2112.13246}, 2021.

\bibitem{timeevolving2}
J.~Dong, D.~Zhang, Y.~Cong, W.~Cong, H.~Ding, and D.~Dai, ``Federated incremental semantic segmentation,'' in \emph{2023 IEEE/CVF Conference on Computer Vision and Pattern Recognition (CVPR)}, 2023.

\bibitem{elastic}
D.~Chen, J.~Hu, V.~J. Tan, X.~Wei, and E.~Wu, ``Elastic aggregation for federated optimization,'' in \emph{Proceedings of the IEEE Conference on Computer Vision and Pattern Recognition (CVPR)}, 2023, pp. 12\,187--12\,197.

\bibitem{fedseg}
J.~Miao, Z.~Yang, L.~Fan, and Y.~Yang, ``Fedseg: Class-heterogeneous federated learning for semantic segmentation,'' in \emph{Proceedings of the IEEE Conference on Computer Vision and Pattern Recognition (CVPR)}, June 2023, pp. 8042--8052.

\bibitem{fedh2l}
Y.~Li, W.~Zhou, H.~Wang, H.~Mi, and T.~M. Hospedales, ``Fedh2l: Federated learning with model and statistical heterogeneity,'' \emph{arXiv preprint arXiv:2101.11296}, 2021.

\bibitem{fedfed}
Z.~Yang, Y.~Zhang, Y.~Zheng, X.~Tian, H.~Peng, T.~Liu, and B.~Han, ``Fedfed: Feature distillation against data heterogeneity in federated learning,'' \emph{Advances in Neural Information Processing Systems}, vol.~36, 2024.

\bibitem{fedotp}
H.~Li, W.~Huang, J.~Wang, and Y.~Shi, ``Global and local prompts cooperation via optimal transport for federated learning,'' in \emph{Proceedings of the IEEE Conference on Computer Vision and Pattern Recognition (CVPR)}, 2024, pp. 12\,151--12\,161.

\bibitem{fedaf}
Y.~Wang, H.~Fu, R.~Kanagavelu, Q.~Wei, Y.~Liu, and R.~S.~M. Goh, ``An aggregation-free federated learning for tackling data heterogeneity,'' in \emph{Proceedings of the IEEE Conference on Computer Vision and Pattern Recognition (CVPR)}, June 2024, pp. 26\,233--26\,242.

\bibitem{cfl}
Y.~Guo, T.~Lin, and X.~Tang, ``Towards federated learning on time-evolving heterogeneous data,'' \emph{ArXiv}, vol. abs/2112.13246, 2021.

\bibitem{fedthe}
L.~Jiang and T.~Lin, ``Test-time robust personalization for federated learning,'' \emph{arXiv preprint arXiv:2205.10920}, 2022.

\bibitem{pfedem}
Y.~Yeganeh, A.~Farshad, J.~Boschmann, R.~Gaus, M.~Frantzen, and N.~Navab, ``Adaptive personlization in federated learning for highly non-i.i.d. data,'' \emph{arXiv preprint arXiv:2207.03448}, 2022.

\bibitem{fedrc}
Y.~Guo, X.~Tang, and T.~Lin, ``Fedrc: Tackling diverse distribution shifts challenge in federated learning by robust clustering,'' \emph{arXiv preprint arXiv:2301.12379}, 2023.

\bibitem{fedpm}
X.~Zhang, B.~Zhang, W.~Yu, and X.~Kang, ``Federated deep learning with prototype matching for object extraction from very-high-resolution remote sensing images,'' \emph{IEEE Transactions on Geoscience and Remote Sensing}, vol.~61, pp. 1--16, 2023.

\bibitem{geofed}
\BIBentryALTinterwordspacing
J.~Tan, Y.~Li, S.~A. Bartalev, B.~Dang, W.~Chen, Y.~Zhang, and L.~Yuan, ``Bridging data islands: Geographic heterogeneity-aware federated learning for collaborative remote sensing semantic segmentation,'' 2024. [Online]. Available: \url{https://arxiv.org/abs/2404.09292}
\BIBentrySTDinterwordspacing

\bibitem{dmcn}
M.~J. Khan, S.~Rath, and M.~H. Zaib, ``Privacy-enhanced image restoration in remote sensing via federated learning,'' in \emph{2024 12th International Symposium on Digital Forensics and Security (ISDFS)}, 2024, pp. 1--6.

\bibitem{adaptive}
P.~Tam, S.~Math, C.~Nam, and S.~Kim, ``Adaptive resource optimized edge federated learning in real-time image sensing classifications,'' \emph{IEEE Journal of Selected Topics in Applied Earth Observations and Remote Sensing}, vol.~14, pp. 10\,929--10\,940, 2021.

\bibitem{uav}
I.~Abunadi, M.~M. Althobaiti, F.~N. Al-Wesabi, A.~M. Hilal, M.~Medani, M.~A. Hamza, M.~Rizwanullah, and A.~S. Zamani, ``Federated learning with blockchain assisted image classification for clustered uav networks,'' \emph{Comput. mater. contin}, vol.~72, pp. 1195--1212, 2022.

\bibitem{ppfl}
J.~Zhu, J.~Wu, A.~K. Bashir, Q.~Pan, and W.~Yang, ``Privacy-preserving federated learning of remote sensing image classification with dishonest majority,'' \emph{IEEE Journal of Selected Topics in Applied Earth Observations and Remote Sensing}, vol.~16, pp. 4685--4698, 2023.

\bibitem{ckks}
J.~H. Cheon, A.~Kim, M.~Kim, and Y.~Song, ``Homomorphic encryption for arithmetic of approximate numbers,'' in \emph{Advances in Cryptology--ASIACRYPT 2017: 23rd International Conference on the Theory and Applications of Cryptology and Information Security, Hong Kong, China, December 3-7, 2017, Proceedings, Part I 23}.\hskip 1em plus 0.5em minus 0.4em\relax Springer, 2017, pp. 409--437.

\bibitem{multimodal}
B.~B{\"u}y{\"u}kta{\c{s}}, G.~Sumbul, and B.~Demir, ``Learning across decentralized multi-modal remote sensing archives with federated learning,'' in \emph{IGARSS 2023-2023 IEEE International Geoscience and Remote Sensing Symposium}.\hskip 1em plus 0.5em minus 0.4em\relax IEEE, 2023, pp. 4966--4969.

\bibitem{leveragingfeature}
A.-K. Duong, H.~Ân Lê, and M.-T. Pham, ``Leveraging feature communication in federated learning for remote sensing image classification,'' \emph{arXiv preprint arXiv:2403.13575}, 2024.

\bibitem{dcgan}
O.~Jockusch, M.~Z. Hossain, A.~Imteaj, and A.~R. Shahid, ``Generative ai-based land cover classification via federated learning cnns: Sustainable insights from uav imagery,'' in \emph{2024 IEEE Conference on Technologies for Sustainability (SusTech)}.\hskip 1em plus 0.5em minus 0.4em\relax IEEE, 2024, pp. 356--361.

\bibitem{transformer}
B.~B{\"u}y{\"u}kta{\c{s}}, K.~Weitzel, S.~V{\"o}lkers, F.~Zailskas, and B.~Demir, ``Transformer-based federated learning for multi-label remote sensing image classification,'' \emph{arXiv preprint arXiv:2405.15405}, 2024.

\bibitem{feddiff}
D.~Li, W.~Xie, Z.~Wang, Y.~Lu, Y.~Li, and L.~Fang, ``Feddiff: Diffusion model driven federated learning for multi-modal and multi-clients,'' \emph{IEEE Transactions on Circuits and Systems for Video Technology}, 2024.

\bibitem{rs-clip}
X.~Li, C.~Wen, Y.~Hu, and N.~Zhou, ``Rs-clip: Zero shot remote sensing scene classification via contrastive vision-language supervision,'' \emph{International Journal of Applied Earth Observation and Geoinformation}, 2023.

\bibitem{changeclip}
S.~Dong, L.~Wang, B.~Du, and X.~Meng, ``Changeclip: Remote sensing change detection with multimodal vision-language representation learning,'' \emph{ISPRS Journal of Photogrammetry and Remote Sensing}, vol. 208, pp. 53--69, 2024.

\bibitem{pir-clip}
J.~Pan, M.~Ma, Q.~Ma, C.~Bai, and S.~Chen, ``Pir: Remote sensing image-text retrieval with prior instruction representation learning,'' \emph{arXiv preprint arXiv:2405.10160}, 2024.

\bibitem{sg-clip}
L.~Liu, L.~Yang, F.~Yang, F.~Chen, and F.~Xu, ``Clip-driven few-shot species-recognition method for integrating geographic information,'' \emph{Remote Sensing}, vol.~16, no.~12, p. 2238, 2024.

\bibitem{geochat}
K.~Kuckreja, M.~S. Danish, M.~Naseer, A.~Das, S.~Khan, and F.~S. Khan, ``Geochat: Grounded large vision-language model for remote sensing,'' in \emph{Proceedings of the IEEE/CVF Conference on Computer Vision and Pattern Recognition}, 2024, pp. 27\,831--27\,840.

\bibitem{llava}
C.~Li, C.~Wong, S.~Zhang, N.~Usuyama, H.~Liu, J.~Yang, T.~Naumann, H.~Poon, and J.~Gao, ``Llava-med: Training a large language-and-vision assistant for biomedicine in one day,'' \emph{Advances in Neural Information Processing Systems}, vol.~36, 2024.

\bibitem{lora}
E.~J. Hu, Y.~Shen, P.~Wallis, Z.~Allen-Zhu, Y.~Li, S.~Wang, L.~Wang, and W.~Chen, ``Lora: Low-rank adaptation of large language models,'' \emph{arXiv preprint arXiv:2106.09685}, 2021.

\bibitem{modalitygap}
A.~Zavras, D.~Michail, B.~Demir, and I.~Papoutsis, ``Mind the modality gap: Towards a remote sensing vision-language model via cross-modal alignment,'' \emph{arXiv preprint arXiv:2402.09816}, 2024.

\bibitem{optimal-31}
Q.~Wang, S.~Liu, J.~Chanussot, and X.~Li, ``Scene classification with recurrent attention of vhr remote sensing images,'' \emph{IEEE Transactions on Geoscience and Remote Sensing}, vol.~57, no.~2, pp. 1155--1167, 2019.

\bibitem{UCMerced}
Y.~Yang and S.~Newsam, ``Bag-of-visual-words and spatial extensions for land-use classification,'' 11 2010, pp. 270--279.

\bibitem{nwpu}
G.~Cheng, J.~Han, and X.~Lu, ``Remote sensing image scene classification: Benchmark and state of the art,'' \emph{Proceedings of the IEEE}, vol. 105, no.~10, pp. 1865--1883, 2017.

\bibitem{chizat2018scaling}
L.~Chizat, G.~Peyr{\'e}, B.~Schmitzer, and F.-X. Vialard, ``Scaling algorithms for unbalanced optimal transport problems,'' \emph{Mathematics of Computation}, vol.~87, no. 314, pp. 2563--2609, 2018.

\bibitem{benamou2015iterative}
J.-D. Benamou, G.~Carlier, M.~Cuturi, L.~Nenna, and G.~Peyr{\'e}, ``Iterative bregman projections for regularized transportation problems,'' \emph{SIAM Journal on Scientific Computing}, vol.~37, no.~2, pp. A1111--A1138, 2015.

\bibitem{dykstra1983algorithm}
R.~L. Dykstra, ``An algorithm for restricted least squares regression,'' \emph{Journal of the American Statistical Association}, vol.~78, no. 384, pp. 837--842, 1983.

\bibitem{ben2024federated}
B.~Ben~Youssef, L.~Alhmidi, Y.~Bazi, and M.~Zuair, ``Federated learning approach for remote sensing scene classification,'' \emph{Remote Sensing}, vol.~16, no.~12, p. 2194, 2024.

\end{thebibliography}
}

\end{document}